\pdfoutput=1

\documentclass[11pt]{article}

 \usepackage[final]{acl}

\usepackage{times}
\usepackage{latexsym}

\usepackage[T1]{fontenc}

\usepackage[utf8]{inputenc}

\usepackage{microtype}

\usepackage{inconsolata}

\usepackage{graphicx}

\usepackage{multirow}
\usepackage{multicol}
\usepackage[normalem]{ulem}
\usepackage{paralist}
\usepackage{booktabs}
\usepackage{comment}
\usepackage{amsmath}
\usepackage{soul}

\usepackage{algorithmic}
\usepackage{algorithm}

\useunder{\uline}{\ul}{}

\title{Selective Self-to-Supervised Fine-Tuning for Generalization in Large Language Models}

\author{Sonam Gupta \thanks{Corresponding author: Sonam.Gupta7@ibm.com}, Yatin Nandwani, Asaf Yehudai, Dinesh Khandelwal,\\ {\bf Dinesh Raghu and Sachindra Joshi} \\ 
          IBM Research \\}
         
\begin{document}
\maketitle
\begin{abstract}
Fine-tuning Large Language Models (LLMs) on specific datasets is a common practice to improve performance on target tasks. However, this performance gain often leads to overfitting, where the model becomes too specialized in either the task or the characteristics of the training data, resulting in a loss of generalization. This paper introduces Selective Self-to-Supervised Fine-Tuning (S3FT), a fine-tuning approach that achieves better performance than the standard supervised fine-tuning (SFT) while improving generalization.
S3FT leverages the existence of multiple valid responses to a query.
By utilizing the model's correct responses, S3FT reduces model specialization during the fine-tuning stage. S3FT first identifies the correct model responses from the training set by deploying an appropriate judge. Then, it fine-tunes the model using the correct model responses and the gold response (or its paraphrase) for the remaining samples.
The effectiveness of S3FT is demonstrated through experiments on mathematical reasoning, Python programming and reading comprehension tasks. The results show that standard SFT can lead to an average performance drop of up to $4.4$ on multiple benchmarks, such as MMLU and TruthfulQA. In contrast, S3FT reduces this drop by half, i.e. $2.5$, indicating better generalization capabilities than SFT while performing significantly better on the fine-tuning tasks.
\end{abstract}
\section{Introduction}
\newcommand{\mistral}{Mistral-7B-Instruct-v0.2}

\begin{table}[t]
\small
\begin{tabular}{p{0.2\columnwidth}@{}|p{0.75\columnwidth}@{}}
\toprule
\textit{Task} & Write a function to add two lists using map and lambda function. \\ \cmidrule(l){1-2}
\textit{Gold} & def add\_list(nums1,nums2): \\
  & \ \ result = map(lambda x, y:x+y, nums1, nums2) \\
  & \ \ return list(result) \\
& \sethlcolor{cyan!30} \hl{log probability = -84.32} \\ \cmidrule(l){1-2}
\textit{Model} & def add\_list(list1, list2):\\
\textit{Prediction} & \ \ return list(map(lambda x, y:x+y, list1, list2)) \\
& \sethlcolor{lime} \hl{log probability = -36.64} \\ \bottomrule
\end{tabular}
\caption{An example from the MBPP (Python Programming dataset) \cite{austin2021program}, along with \mistral's prediction.}
\label{tab:intro-example}
\vspace{-0.5em}
\end{table}

Large Language Models (LLMs) have made remarkable progress in recent years, demonstrating impressive capabilities across a wide range of tasks, including question-answering \cite{rajpurkar2016squad}, summarization \cite{nallapati2016abstractive}, and more \cite{brown2020language}. Supervised fine-tuning (SFT) of LLMs on task-specific data is a widely used approach to enhance their performance in specialized applications. While fine-tuning improves task accuracy, it can cause the model to overfit to the domain or style present in the training data, potentially limiting its broader applicability. In this work, we focus on exploring how to fine-tune an LLM for a specific task while preserving its general-purpose capabilities.

SFT relies on gold responses for training. We make two key observations when performing SFT on LLMs: (1) in many datasets, model-generated responses—though differing from gold responses—are still valid and acceptable, and (2) the distribution of gold responses often diverges significantly from the model’s own response distribution. For instance, consider an example from the MBPP dataset in Table \ref{tab:intro-example}. The base model, \mistral, assigns a log probability of $-84.32$ to the gold answer. When prompted with the same question, the model generates a response containing the same information as the gold output but with a much higher log probability of $-36.64$. This phenomenon is common in generation tasks, where semantically equivalent responses can have widely varying log-likelihood scores. This observation suggests that model-generated responses can align more closely with the model's native distribution, whereas gold responses may lie further apart. As a result, training exclusively on gold responses risks introducing distributional drift, potentially reducing the model’s ability to generalize effectively.

To address this issue, we propose Selective Self-to-Supervised Fine-Tuning (S3FT), a simple yet powerful technique that utilizes model-generated answers for a subset of the training dataset to adapt the model to desirable behaviours while maintaining generalization. S3FT fine-tunes the model on its own generated output for cases where it behaves desirably and on gold output (or its paraphrase) for the remaining data. This approach allows the model to learn from its successes while benefiting from human-labeled data when needed.

In our experiments, we show that S3FT outperform existing approaches, including SFT, on diverse tasks, namely code generation~\citet{austin2021program}, math problem solving~\cite{cobbe2021training} and reading comprehension~\cite{kwiatkowski2019natural}. To show that S3FT generalizes better and retains the base model's capabilities, we evaluate on multiple datasets such as MMLU \cite{mmlu}, TruthfulQA \cite{truthfulqa}, and Hellaswag \cite{hellaswag}. We observe that the drop in performance for S3FT on these benchmarks is smaller than that of existing approaches, demonstrating better generalization capabilities.

\section{Proposed Method}
\newcommand{\model}{\mathcal{M}}
\newcommand{\basemodel}{\mathcal{M}_{\theta_0}}

\newcommand{\task}{\mathcal{T}}
\newcommand{\traindata}{\mathcal{D}}
\newcommand{\inp}{x}
\newcommand{\outp}{y}
\newcommand{\loss}{\mathcal{L}}
\newcommand{\sftloss}{\mathcal{L}_{SFT}}
\newcommand{\prob}{Pr}
\newcommand{\pred}{\hat{y}}
\newcommand{\etc}{\emph{etc.}}
\newcommand{\goodsubset}{\mathcal{R}}
\newcommand{\goldsubset}{\mathcal{G}}

\newcommand{\ssrloss}{\mathcal{L}_{SSR}}

\begin{figure*}
    \centering
    \includegraphics[width=\linewidth]{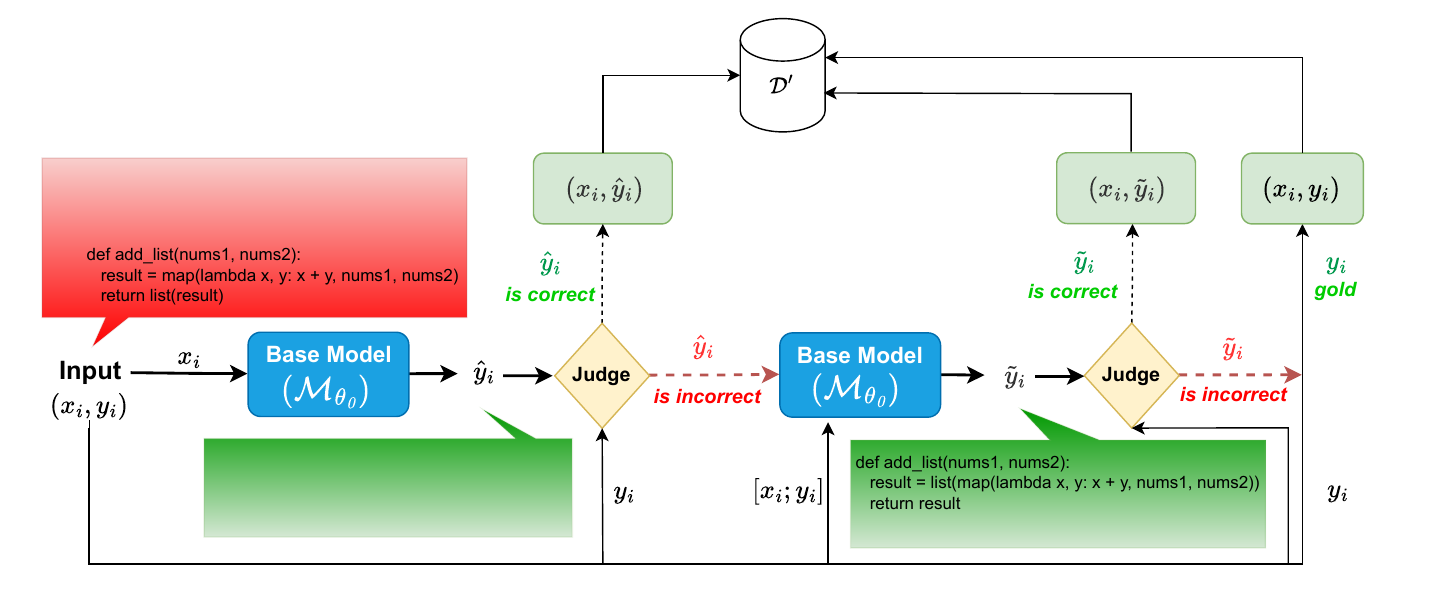}
    \caption{\textbf{An overview of our proposed approach}: Given the input $\inp_i$ and its corresponding gold response $\outp_i$, we employ the base model $\mathcal{M}_{\theta}$ to transform $\outp_i$ such that it is correct but at the same time closer to model's distribution. First, the model predicts the output $\hat{y}$. The judge decides whether the $\outp_i$ is correct. If true, it becomes part of the training dataset; otherwise, we paraphrase $([x_i;y_i])$ to obtain $\tilde{y_i}$. The judge evaluates the correctness of $\tilde{y_i}$. If true, we use $\tilde{y_i}$; otherwise, we use $\outp_i$ as the target response. The resulting dataset $\mathcal{D'}$ is used to train the model.}
    \label{fig:overview}
    \vspace{-0.5em}
\end{figure*}

Let $\model_\theta$ parameterized by $\theta$ be a given large language model. 
Let $\theta = \theta_0$ be the given model weights obtained after pre-training and instruction tuning the model.
We refer to $\model_{\theta_0}$ as the base model.
Further, let $\task$ be the new task that we wish to teach the model $\model_\theta$, and let $\traindata = \{(\inp_i, \outp_i)\}_{i = 1}^{N}$ be the corresponding training dataset for $\task$.
In standard SFT, we backpropagate through the standard Cross Entropy loss over the training dataset, $\traindata$.
However, SFT can cause a degradation of $\model_{\theta_0}$ general capabilities by forcing the model to predict a gold response which is further away from the $\mathcal{M}_{\theta_0}$ responses' distribution.

To solve this, our method relies on two key observations.
First, for many NLP tasks e.g. machine translation, summarization, reading comprehension \etc, the same input $\inp$ can have multiple
valid responses. \citet{nandwani2020neural} define this setup as 1oML (one of many learning) and propose various strategies to handle it, albeit for combinatorial problems.
Second, teaching the model using its own words can help preserve its own distribution. This can regularizes the model training, helping it to tackle catastrophic forgetting and maintaining its general capabilities.
We note that the standard practice of regularization via replay buffer \cite{hayes2020remind},
which involves mixing a subset of instruction-tuning dataset with the given task-specific data $\traindata$ is not always feasible as the instruction-tuning dataset may not be available.
These two observations are the basis for S3FT.
For each example in the data, we start by generating a prediction $\pred_i = \model_{\theta_0}(\inp_i)$ with the base model where $\inp_i$ is the input. 
If $\pred_i$ is equivalent to $\outp_i$, we use $(\inp_i$, $\pred_i)$ for model training. If $\pred_i$ is not equivalent, we use $\model_{\theta_0}$ to rephrase the gold answer $\outp_i$ in its own language, $\tilde{y_i} = \model_{\theta_0}([\inp_i; \outp_i])$. If $\tilde{y_i}$ is not equivalent to $\outp_i$ we use the gold answer, $\outp_i$. Figure \ref{fig:overview} and the algorithm in Appendix \ref{app:algo} gives an overview of our approach.

An important component of S3FT is the ability to identify the equivalence of model's prediction or gold paraphrasing to the gold answer.
Towards that end, we can use judges that aim to assess the equivalence with $\pred_i$, either by heuristics such as checking the bottom-line agreement of the predicted and gold response or employing a stronger LLM as a judge to measure more semantic equivalence.

\section{Experimental Setup}
\noindent We evaluate and compare the proposed S3FT method with the vanilla SFT and other methods that try to perform fine-tuning while retaining generalization capabilities. We aim to answer the following research questions: \textbf{RQ1. In-Domain Performance:} How well does S3FT learn the fine-tuning task compared to baselines when fine-tuned and evaluated on the same dataset? \textbf{RQ2. Generalization:} How well does S3FT retain the inherent capabilities of the base model post fine-tuning? \textbf{RQ3. Effect of gold response's paraphrasing:} How beneficial is it to use gold paraphrases that are closer to the base model's distribution? 

\noindent \textbf{Datasets:} 
We focus on three tasks, improving the mathematical reasoning abilities, basic python programming and reading comprehension skills. For enhancing the mathematical skills we experiment with GSM8K~\cite{cobbe2021training} dataset. This dataset consists of grade school level math word problems and solutions. To boost the Python programming skill, we experiment with MBPP~\cite{austin2021program} dataset which consists of a task description, three test cases, and a code solution for each example. We experiment with a variant of NQ~\cite{kwiatkowski2019natural} dataset as introduced in \cite{slobodkin2023curious} to enhance the reading-comprehension skills. The NQ dataset from \cite{slobodkin2023curious}  contains 3800 (context, question, answer) pairs. The questions are categorized into two types: (i) answerable -- where the provided context is relevant and contains sufficient information to derive an answer, and ii) unanswerable -- where the context lacks the necessary information to answer the question. A detailed description of the datasets, along with the training, validation and testing splits, is provided in the Appendix \ref{app:data_detail}.
\vspace{0.2ex}

\noindent \textbf{Evaluation Metrics:} For GSM8K, we assess the correctness of a generated response by checking if its predicted answer matches the final answer in the gold solution. For MBPP, we evaluate the generated code by executing it against the provided test cases. If the code passes all the test cases, it is considered correct. For NQ dataset, we employ Mistral-instruct-v2 (7B) as a judge. Following \cite{badshah2408reference}, we use a reference-guided prompt to judge the correctness of the generated responses. More details about the LLM judge can be found in appendix \ref{app:llm-judge}. Finally, we report the accuracy of each method for the three datasets.

\noindent \textbf{Human Study on Judges’ Accuracy:} Since responses in GSM8K and NQ are more open-ended, we conduct a human study to assess the reliability of the judges used for evaluating correctness. For GSM8K, we randomly sampled 50 examples and found that the judge is approximately 96\% accurate. Similarly, for NQ, testing on a set of 200 random samples yielded an accuracy of 86\%. Additionally, we highlight that improving the quality of the judges could further enhance the evaluation process and improve the overall generalizability of the method.

\vspace{0.2ex}
\noindent \textbf{Base model and Baselines:} 
We experiment with Mistral-instruct-v2 (7B) \cite{mistral} as our base model. We use three baselines: (1) prompting the base model (see Appendix \ref{app:task-prompts} for the exact prompt), (2) Supervised Fine-Tuning (SFT), and (3) Self-distill Fine-tuning (SDFT) \cite{yang2024self}. SDFT is a contemporary work that adopts gold answer rephrasing to preserve the generalization of the fine-tuned model.

\noindent \textbf{Implementation Details:} For all fine-tuning, we use Low-Rank Adaptation (LoRA) \cite{hu2022lora} with a rank of 8, a scaling factor of 16 and a dropout of 0.1. Please see Appendix \ref{app:train_detail} for more details.

\section{Results and Discussion}
\label{sec:results}
\paragraph{In-Domain Performance}
\label{subsec:same_task_same_domain}
\begin{table}
\centering
\resizebox{0.85\columnwidth}{!}{
\begin{tabular}{@{}lccc@{}} 
\toprule
\textbf{Method}           & \textbf{GSM8K} & \textbf{MBPP} & \textbf{NQ} \\ \midrule
Base & 40.3          & 22.2          & 64.7       \\ 
SFT  & 53.4          & 32.8          & 60.0       \\ 
SDFT & 54.8          & 35.8          & \textbf{67.1}        \\ 
S3FT & \textbf{56.9} & \textbf{39.4} & \textbf{67.1}        \\ 
\bottomrule
\end{tabular}
}
\caption{Performance comparison of various fine tuning techniques over two different tasks using Accuracy(\%) as the metric. S3FT achieves the best performance on the fine-tuning task while preserving the generalization to the other tasks.}
\label{tab:in_domain_recall_acc}
\end{table}

Table \ref{tab:in_domain_recall_acc} reports the accuracy of Mistral-Instruct-v2 (7B) fine-tuned over GSM8K, MBPP and NQ datasets. Here, we evaluate the base, SFT, SDFT and S3FT models over the test set corresponding to the training dataset.

The base model shows suboptimal performance on all the datasets.
Fine-tuning using our method significantly improves the model's performance. S3FT achieves a 2.1\% gain on GSM8K, 3.6\% gain on MBPP and comparable performance to the state-of-the-art method SDFT on reading comprehension. We note that S3FT significantly outperforms SFT on all three datasets. This improvement stems from a simple yet effective technique of using the base model's responses for training when the base model is correct instead of the gold response. This verifies that S3FT learns the fine-tuning task well outperforming all other baselines (RQ1). 
\vspace{0.2ex}
\paragraph{Generalization}

\begin{table*}
\centering
\begin{tabular}{@{}cc|cccc|c@{}}
\toprule
\textbf{\begin{tabular}[c]{@{}c@{}}Train\\ Dataset\end{tabular}} & \textbf{Method} & \textbf{MMLU} & \textbf{\begin{tabular}[c]{@{}c@{}}TruthfulQA\end{tabular}} & \textbf{\begin{tabular}[c]{@{}c@{}}HellaSwag\end{tabular}} & \textbf{\begin{tabular}[c]{@{}c@{}}WinoGrande\end{tabular}} & \textbf{Average} \\ \midrule
& Base & 58.7 & 59.6 & 66.0 & 74.0 & 64.6 \\ \midrule
\multirow{3}{*}{GSM8K} & SFT & 57.0 & 48.0 & 62.4 & 73.4 & 60.2 \\
& SDFT & 57.6 & 51.1 & 62.6 & \textbf{73.6} & 61.2 \\
& S3FT & \textbf{58.2} & \textbf{53.7} & \textbf{63.2} & 73.5 & \textbf{62.1} \\ \midrule
\multirow{3}{*}{MBPP} & SFT & 57.5 & 51.9 & 64.7 & 73.6 & 61.9 \\
& SDFT & 57.1 & 56.6 & \textbf{65.2} & 73.4 & 63.1 \\
& S3FT & \textbf{58.0} & \textbf{57.2} & 64.9 & \textbf{74.4} & \textbf{63.6} \\ \midrule
\multirow{3}{*}{NQ} & SFT & 54.4 & 45.7 & 63.1 & 72.0 & 58.8 \\
& SDFT & 55.2 & 53.9 & \textbf{65.3} & \textbf{72.6} & 61.8 \\
& S3FT & \textbf{57.0} & \textbf{54.1} & 64.8 & \textbf{72.6} & \textbf{62.1} \\ \bottomrule
\end{tabular}
\caption{Generalization over other benchmarks. 1st row reports the score obtained by prompting the base model Mistral-instruct-v2-7B.}
\label{tab:other_tasks_mistral}
\end{table*}
A major issue with SFT is its tendency to diminish the model's previously learned skills.
To demonstrate that S3FT alleviates this issue, we 
compare the baselines and our method against the base model on a diverse set of publicly available benchmarks. Specifically, we evaluate them on MMLU \cite{mmlu}, Truthful-QA \cite{truthfulqa}, Hellaswag \cite{hellaswag} and, Winogrande \cite{sakaguchi2019winogrande}. Table \ref{tab:other_tasks_mistral} reports our findings. 

\begin{figure}
    \centering
    \includegraphics[width=\linewidth]{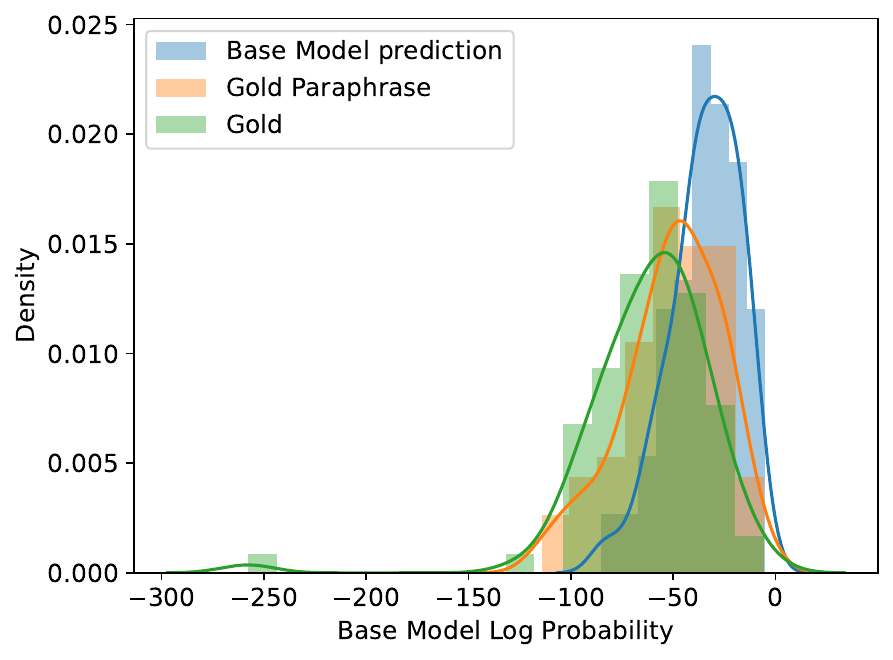}
    \caption{Histogram of the log probability assigned by \mistral\ to the gold responses, paraphrase of gold responses and model's own predictions. The distribution is based on 84 examples from the MBPP training data.}
    \label{fig:log_prob_dist}
    \vspace{-1em}
\end{figure}

We observe that irrespective of the task used for fine-tuning, there is a drop in the performance of SFT models across all benchmarks. 
On average, SFT on Mistral-7B results in an average drop of $4.4$, $2.7$ and $5.8$ when trained using GSM8K, MBPP and NQ respectively. 
On the other hand, S3FT results in an average drop of only $2.5$ when trained on GSM8K and MBPP and a drop of $1.0$ when trained on MBPP.
This clearly demonstrates that our proposed technique for fine-tuning preserves the base model's capabilities (RQ2) without relying upon any kind of replay buffer which might not even be available in many cases. On the other hand, standard SFT results in overfitting to the training dataset, resulting in catastrophic forgetting of the skills acquired by the base model during pre-training and instruction tuning. 
\paragraph{Effect of Gold response's paraphrasing}
Our findings from Tables \ref{tab:in_domain_recall_acc} and \ref{tab:other_tasks_mistral} indicate that using the model's response as the target rather than the gold response significantly enhances fine-tuning performance without compromising the model's overall capabilities. Training on gold responses can cause a shift from the original distribution, negatively impacting the model's generalization. Figure \ref{fig:log_prob_dist} illustrates this gap between the distributions of gold responses, gold paraphrases, and the base model's responses. The distribution is plotted using $84$ random samples for which the model's prediction and the paraphrases of the gold responses are acceptable. The plot shows that in several cases model responses can be valid and closer to model's own distribution and therefore curating the training data in this way leads to minimal changes in the model parameters while fine-tuning. Thus, the closer the gold response's paraphrases to the model distribution, the better it is (RQ3). 

\paragraph{Training Data Proportions}
Table \ref{tab:human_judge_study_proportions} presents the proportions of examples in which S3FT utilizes base model responses, gold paraphrases, and gold responses across the three datasets. Notably, the base model's responses were deemed acceptable for at least 30\% of the training samples. When incorporating gold paraphrases, more than 50\% of the training data originates from the model's own responses. This suggests that the base model's outputs play a crucial role in S3FT's success by acting as an effective regularizer, helping to mitigate overfitting in the fine-tuning process.

\begin{table}
\centering
\resizebox{0.95\columnwidth}{!}{
\begin{tabular}{lccc}
\hline
\multicolumn{1}{c}{\textbf{Dataset}} & \textbf{\begin{tabular}[c]{@{}c@{}}Model \\ responses\end{tabular}} & \textbf{\begin{tabular}[c]{@{}c@{}}Gold \\ paraphrases\end{tabular}} & \textbf{\begin{tabular}[c]{@{}c@{}}Gold \\ responses\end{tabular}} \\
\hline
GSM8K & 49.5\% & 42.1\% & 8.4\% \\
MBPP & 30.2\% & 32.4\% & 37.4\% \\
NQ & 66.0\% & 25.6\% & 8.1\% \\
\hline
\end{tabular}
}
 \caption{The proportions of examples used from each data type; Model responses, Gold paraphrases, and Gold responses. We can see that the exact composition depend on the dataset, but the majority of responses are either the model responses or its paraphrasing.}
 \label{tab:human_judge_study_proportions}
\end{table}

\section{Related Work}
Continual learning for language models faces challenges like overfitting and loss of generalization  \cite{yogatama2019learning,zhang2021revisiting}. Rehearsal-based methods, such as experience replay \cite{rolnick2019experience} and representation consolidation \cite{bhat2022representation}, show promise but often depend on real data, which may be scarce. To address this, model-generated responses are used through techniques like self-training \cite{he2020revisiting,xie2020self} and self-supervised learning \cite{lewis2020bart}. However, their effectiveness in continual learning remains underexplored. Current methods focus on real data rehearsal \cite{scialom2022continual,mok2023large,zhang2023continual}, but these can be resource-intensive. In contrast, S3FT avoids storing past data or training extra generative models, making it more data-efficient and practical for real-world use.
SDFT \cite{yang2024self}, the closest contemporary work to ours, is thoroughly compared in experiments, where we achieve significantly higher accuracy.
\section{Conclusion}
In this paper, we present S3FT, a fine-tuning approach that enhances both task-specific performance and generalization across tasks, as shown on benchmarks like MMLU and Truthful QA. S3FT leverages the idea that multiple correct outputs may exist and avoids unnecessary changes by fine-tuning on gold response (or its paraphrase) only when the model's response is incorrect. In future work, we plan to investigate techniques like few-shot prompting for sampling correct outputs that are closer to the model's own distribution to reduce the changes from the base model's weights.
\section{Limitations}

S3FT requires running model inference on the entire training dataset, identifying correct and incorrect responses, then performing gold rephrasing on incorrect responses and evaluating their correctness. These additional steps introduce a few extra requirements not present in standard SFT.
First, S3FT's improved results come at the cost of increased computational demands. Second, it requires a qualified judge. As we show, reliable heuristics can be used to assess response equivalence in mathematical reasoning and Python programming tasks. However, for open-ended tasks such as summarization and translation, no simple heuristics exist.
Here, we demonstrate that for reading comprehension, an LLM judge can serve as a reliable evaluator. As LLM judges improve, the usability and applicability of our method can extend to a broader range of tasks.

\bibliography{custom}

\begin{thebibliography}{29}
\providecommand{\natexlab}[1]{#1}

\bibitem[{Austin et~al.(2021)Austin, Odena, Nye, Bosma, Michalewski, Dohan,
  Jiang, Cai, Terry, Le et~al.}]{austin2021program}
Jacob Austin, Augustus Odena, Maxwell Nye, Maarten Bosma, Henryk Michalewski,
  David Dohan, Ellen Jiang, Carrie Cai, Michael Terry, Quoc Le, et~al. 2021.
\newblock Program synthesis with large language models.
\newblock \emph{arXiv preprint arXiv:2108.07732}.

\bibitem[{Badshah and Sajjad(2024)}]{badshah2408reference}
Sher Badshah and Hassan Sajjad. 2024.
\newblock Reference-guided verdict: Llms-as-judges in automatic evaluation of
  free-form text, 2024.
\newblock \emph{URL https://arxiv. org/abs/2408.09235}.

\bibitem[{Ben~Allal et~al.(2022)Ben~Allal, Muennighoff, Kumar~Umapathi, Lipkin,
  and von Werra}]{bigcode-evaluation-harness}
Loubna Ben~Allal, Niklas Muennighoff, Logesh Kumar~Umapathi, Ben Lipkin, and
  Leandro von Werra. 2022.
\newblock A framework for the evaluation of code generation models.
\newblock \url{https://github.com/bigcode-project/bigcode-evaluation-harness}.

\bibitem[{Bhat et~al.(2022)Bhat, Sidorov, Paquet, and
  Garg}]{bhat2022representation}
Sarthak Bhat, Oleg Sidorov, Ulrich Paquet, and Anirudh Garg. 2022.
\newblock Representation consolidation for continual learning.
\newblock In \emph{International Conference on Learning Representations}.

\bibitem[{Brown et~al.(2020)Brown, Mann, Ryder, Subbiah, Kaplan, Dhariwal,
  Neelakantan, Shyam, Sastry, Askell et~al.}]{brown2020language}
Tom Brown, Benjamin Mann, Nick Ryder, Melanie Subbiah, Jared~D Kaplan, Prafulla
  Dhariwal, Arvind Neelakantan, Pranav Shyam, Girish Sastry, Amanda Askell,
  et~al. 2020.
\newblock Language models are few-shot learners.
\newblock \emph{Advances in neural information processing systems},
  33:1877--1901.

\bibitem[{Cobbe et~al.(2021)Cobbe, Kosaraju, Bavarian, Chen, Jun, Kaiser,
  Plappert, Tworek, Hilton, Nakano et~al.}]{cobbe2021training}
Karl Cobbe, Vineet Kosaraju, Mohammad Bavarian, Mark Chen, Heewoo Jun, Lukasz
  Kaiser, Matthias Plappert, Jerry Tworek, Jacob Hilton, Reiichiro Nakano,
  et~al. 2021.
\newblock Training verifiers to solve math word problems.
\newblock \emph{arXiv preprint arXiv:2110.14168}.

\bibitem[{Gao et~al.(2024)Gao, Tow, Abbasi, Biderman, Black, DiPofi, Foster,
  Golding, Hsu, Le~Noac'h, Li, McDonell, Muennighoff, Ociepa, Phang, Reynolds,
  Schoelkopf, Skowron, Sutawika, Tang, Thite, Wang, Wang, and
  Zou}]{eval-harness}
Leo Gao, Jonathan Tow, Baber Abbasi, Stella Biderman, Sid Black, Anthony
  DiPofi, Charles Foster, Laurence Golding, Jeffrey Hsu, Alain Le~Noac'h,
  Haonan Li, Kyle McDonell, Niklas Muennighoff, Chris Ociepa, Jason Phang,
  Laria Reynolds, Hailey Schoelkopf, Aviya Skowron, Lintang Sutawika, Eric
  Tang, Anish Thite, Ben Wang, Kevin Wang, and Andy Zou. 2024.
\newblock \href {https://doi.org/10.5281/zenodo.12608602} {A framework for
  few-shot language model evaluation}.

\bibitem[{Hayes et~al.(2020)Hayes, Kafle, Shrestha, Acharya, and
  Kanan}]{hayes2020remind}
Tyler~L Hayes, Kushal Kafle, Robik Shrestha, Manoj Acharya, and Christopher
  Kanan. 2020.
\newblock Remind your neural network to prevent catastrophic forgetting.
\newblock In \emph{European Conference on Computer Vision}, pages 466--483.
  Springer.

\bibitem[{He et~al.(2020)He, Gu, Shen, and Ranzato}]{he2020revisiting}
Junxian He, Jiatao Gu, Jianfeng Shen, and Marc'Aurelio Ranzato. 2020.
\newblock Revisiting self-training for neural sequence generation.
\newblock In \emph{International Conference on Learning Representations}.

\bibitem[{Hendrycks et~al.(2020)Hendrycks, Burns, Basart, Zou, Mazeika, Song,
  and Steinhardt}]{mmlu}
Dan Hendrycks, Collin Burns, Steven Basart, Andy Zou, Mantas Mazeika, Dawn
  Song, and Jacob Steinhardt. 2020.
\newblock Measuring massive multitask language understanding.
\newblock In \emph{International Conference on Learning Representations}.

\bibitem[{Hu et~al.(2022)Hu, Shen, Wallis, Allen-Zhu, Li, Wang, Wang, and
  Chen}]{hu2022lora}
Edward~J Hu, Yelong Shen, Phillip Wallis, Zeyuan Allen-Zhu, Yuanzhi Li, Shean
  Wang, Lu~Wang, and Weizhu Chen. 2022.
\newblock \href {https://openreview.net/forum?id=nZeVKeeFYf9} {Lo{RA}: Low-rank
  adaptation of large language models}.
\newblock In \emph{International Conference on Learning Representations}.

\bibitem[{Jiang et~al.(2023)Jiang, Sablayrolles, Mensch, Bamford, Chaplot,
  Casas, Bressand, Lengyel, Lample, Saulnier et~al.}]{mistral}
Albert~Q Jiang, Alexandre Sablayrolles, Arthur Mensch, Chris Bamford,
  Devendra~Singh Chaplot, Diego de~las Casas, Florian Bressand, Gianna Lengyel,
  Guillaume Lample, Lucile Saulnier, et~al. 2023.
\newblock Mistral 7b.
\newblock \emph{arXiv preprint arXiv:2310.06825}.

\bibitem[{Kwiatkowski et~al.(2019)Kwiatkowski, Palomaki, Redfield, Collins,
  Parikh, Alberti, Epstein, Polosukhin, Devlin, Lee
  et~al.}]{kwiatkowski2019natural}
Tom Kwiatkowski, Jennimaria Palomaki, Olivia Redfield, Michael Collins, Ankur
  Parikh, Chris Alberti, Danielle Epstein, Illia Polosukhin, Jacob Devlin,
  Kenton Lee, et~al. 2019.
\newblock Natural questions: a benchmark for question answering research.
\newblock \emph{Transactions of the Association for Computational Linguistics},
  7:453--466.

\bibitem[{Lewis et~al.(2020)Lewis, Liu, Goyal, Ghazvininejad, Mohamed, Levy,
  Stoyanov, and Zettlemoyer}]{lewis2020bart}
Mike Lewis, Yinhan Liu, Naman Goyal, Marjan Ghazvininejad, Abdelrahman Mohamed,
  Omer Levy, Ves Stoyanov, and Luke Zettlemoyer. 2020.
\newblock Bart: Denoising sequence-to-sequence pre-training for natural
  language generation, translation, and comprehension.
\newblock In \emph{Proceedings of the 58th Annual Meeting of the Association
  for Computational Linguistics}, pages 7871--7880.

\bibitem[{Lin et~al.(2022)Lin, Hilton, and Evans}]{truthfulqa}
Stephanie Lin, Jacob Hilton, and Owain Evans. 2022.
\newblock {TruthfulQA}: Measuring how models mimic human falsehoods.
\newblock In \emph{Proceedings of the 60th Annual Meeting of the Association
  for Computational Linguistics (Volume 1: Long Papers)}, pages 3214--3252.

\bibitem[{Mok et~al.(2023)Mok, Wellhausen, Choe, and Hajishirzi}]{mok2023large}
Tanya Mok, Luisa Wellhausen, Hyung~Won Choe, and Hannaneh Hajishirzi. 2023.
\newblock Large language models can be continuously updated without forgetting.
\newblock \emph{arXiv preprint arXiv:2303.01926}.

\bibitem[{Nallapati et~al.(2016)Nallapati, Zhou, dos Santos, Gul{\c{c}}ehre,
  and Xiang}]{nallapati2016abstractive}
Ramesh Nallapati, Bowen Zhou, Cicero dos Santos, {\c{C}}aglar Gul{\c{c}}ehre,
  and Bing Xiang. 2016.
\newblock Abstractive text summarization using sequence-to-sequence rnns and
  beyond.
\newblock In \emph{Conference on Computational Natural Language Learning}.
  Association for Computational Linguistics (ACL).

\bibitem[{Nandwani et~al.(2020)Nandwani, Jindal, Singla
  et~al.}]{nandwani2020neural}
Yatin Nandwani, Deepanshu Jindal, Parag Singla, et~al. 2020.
\newblock Neural learning of one-of-many solutions for combinatorial problems
  in structured output spaces.
\newblock In \emph{International Conference on Learning Representations}.

\bibitem[{Rajpurkar et~al.(2016)Rajpurkar, Zhang, Lopyrev, and
  Liang}]{rajpurkar2016squad}
Pranav Rajpurkar, Jian Zhang, Konstantin Lopyrev, and Percy Liang. 2016.
\newblock Squad: 100,000+ questions for machine comprehension of text.
\newblock In \emph{Proceedings of the 2016 Conference on Empirical Methods in
  Natural Language Processing}, pages 2383--2392.

\bibitem[{Rolnick et~al.(2019)Rolnick, Ahuja, Schwarz, Lillicrap, and
  Wayne}]{rolnick2019experience}
David Rolnick, Arun Ahuja, Jonathan Schwarz, Timothy Lillicrap, and Gregory
  Wayne. 2019.
\newblock Experience replay for continual learning.
\newblock In \emph{Advances in Neural Information Processing Systems},
  volume~32.

\bibitem[{Sakaguchi et~al.(2019)Sakaguchi, Bras, Bhagavatula, and
  Choi}]{sakaguchi2019winogrande}
Keisuke Sakaguchi, Ronan~Le Bras, Chandra Bhagavatula, and Yejin Choi. 2019.
\newblock Winogrande: An adversarial winograd schema challenge at scale.
\newblock \emph{arXiv preprint arXiv:1907.10641}.

\bibitem[{Scialom et~al.(2022)Scialom, Charnois, and
  Lamprier}]{scialom2022continual}
Thomas Scialom, Thierry Charnois, and Sylvain Lamprier. 2022.
\newblock Continual learning for large language models.
\newblock In \emph{Findings of the Association for Computational Linguistics:
  EMNLP 2022}, pages 5432--5442.

\bibitem[{Slobodkin et~al.(2023)Slobodkin, Goldman, Caciularu, Dagan, and
  Ravfogel}]{slobodkin2023curious}
Aviv Slobodkin, Omer Goldman, Avi Caciularu, Ido Dagan, and Shauli Ravfogel.
  2023.
\newblock The curious case of hallucinatory (un) answerability: Finding truths
  in the hidden states of over-confident large language models.
\newblock In \emph{Proceedings of the 2023 Conference on Empirical Methods in
  Natural Language Processing}, pages 3607--3625.

\bibitem[{Xie et~al.(2020)Xie, Luong, Hovy, and Le}]{xie2020self}
Qizhe Xie, Minh-Thang Luong, Eduard Hovy, and Quoc~V Le. 2020.
\newblock Self-training with noisy student improves imagenet classification.
\newblock In \emph{Proceedings of the IEEE/CVF Conference on Computer Vision
  and Pattern Recognition}, pages 10687--10698.

\bibitem[{Yang et~al.(2024)Yang, Pang, Feng, Wang, Chen, Zhu, and
  Liu}]{yang2024self}
Zhaorui Yang, Tianyu Pang, Haozhe Feng, Han Wang, Wei Chen, Minfeng Zhu, and
  Qian Liu. 2024.
\newblock Self-distillation bridges distribution gap in language model
  fine-tuning.
\newblock \emph{arXiv preprint arXiv:2402.13669}.

\bibitem[{Yogatama et~al.(2019)Yogatama, d'Autume, Connor, Kocisky,
  Chrzanowski, Kong, Lazaridou, Ling, Yu, Dyer et~al.}]{yogatama2019learning}
Dani Yogatama, Cyprien de~Masson d'Autume, Jerome Connor, Tomas Kocisky, Mike
  Chrzanowski, Lingpeng Kong, Angeliki Lazaridou, Wang Ling, Lei Yu, Chris
  Dyer, et~al. 2019.
\newblock Learning and evaluating general linguistic intelligence.
\newblock \emph{arXiv preprint arXiv:1901.11373}.

\bibitem[{Zellers et~al.(2019)Zellers, Holtzman, Bisk, Farhadi, and
  Choi}]{hellaswag}
Rowan Zellers, Ari Holtzman, Yonatan Bisk, Ali Farhadi, and Yejin Choi. 2019.
\newblock Hellaswag: Can a machine really finish your sentence?
\newblock In \emph{Proceedings of the 57th Annual Meeting of the Association
  for Computational Linguistics}, pages 4791--4800.

\bibitem[{Zhang et~al.(2021)Zhang, Kishore, Wu, Weinberger, and
  Artzi}]{zhang2021revisiting}
Tianyi Zhang, Varsha Kishore, Felix Wu, Kilian~Q Weinberger, and Yoav Artzi.
  2021.
\newblock Bertscore: Evaluating text generation with bert.
\newblock In \emph{International Conference on Learning Representations}.

\bibitem[{Zhang et~al.(2023)Zhang, Wu, Guan, Chen, and
  Zhang}]{zhang2023continual}
Zijun Zhang, Yue Wu, Hao Guan, Xinlei Chen, and Yue Zhang. 2023.
\newblock Continual learning with transformers: Challenges and solutions.
\newblock \emph{arXiv preprint arXiv:2302.13713}.

\end{thebibliography}

\clearpage
\pagebreak
\appendix
\section{Appendix}
\label{sec:appendix}
\subsection{Prompt for Generating the Response}
\label{app:task-prompts}
\begin{figure}
    \centering
    \includegraphics[width=\linewidth]{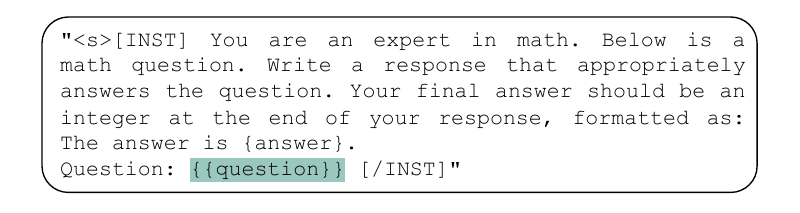}
    \caption{Prompt used for training the model on GSM8K dataset.}
    \vspace{-0.5em}
    \label{fig:gsm8k_train_prompt}
\end{figure}

\begin{figure}
    \centering
    \includegraphics[width=\linewidth]{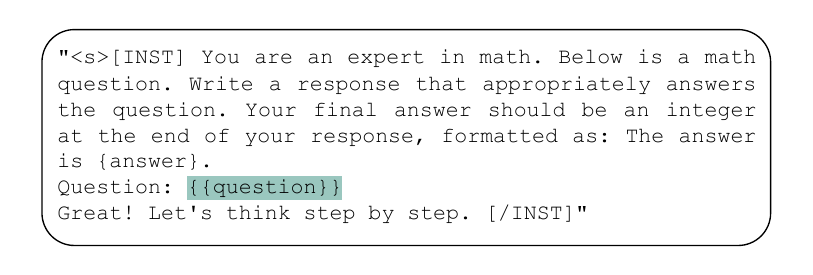}
    \caption{Prompt used for predicting the base model's output on GSM8K dataset.}
    \vspace{-0.5em}
    \label{fig:gsm8k_base_res_prompt}
\end{figure}

\begin{figure}
    \centering
    \includegraphics[width=\linewidth]{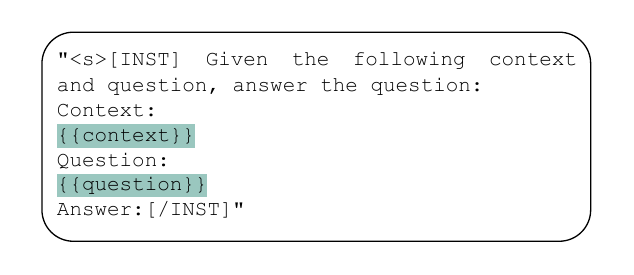}
    \caption{Prompt used for predicting the base model's output on NQ dataset.}
    \vspace{-0.5em}
    \label{fig:nq_base_prompt}
\end{figure}

\begin{figure*}
    \centering
    \includegraphics[width=\linewidth]{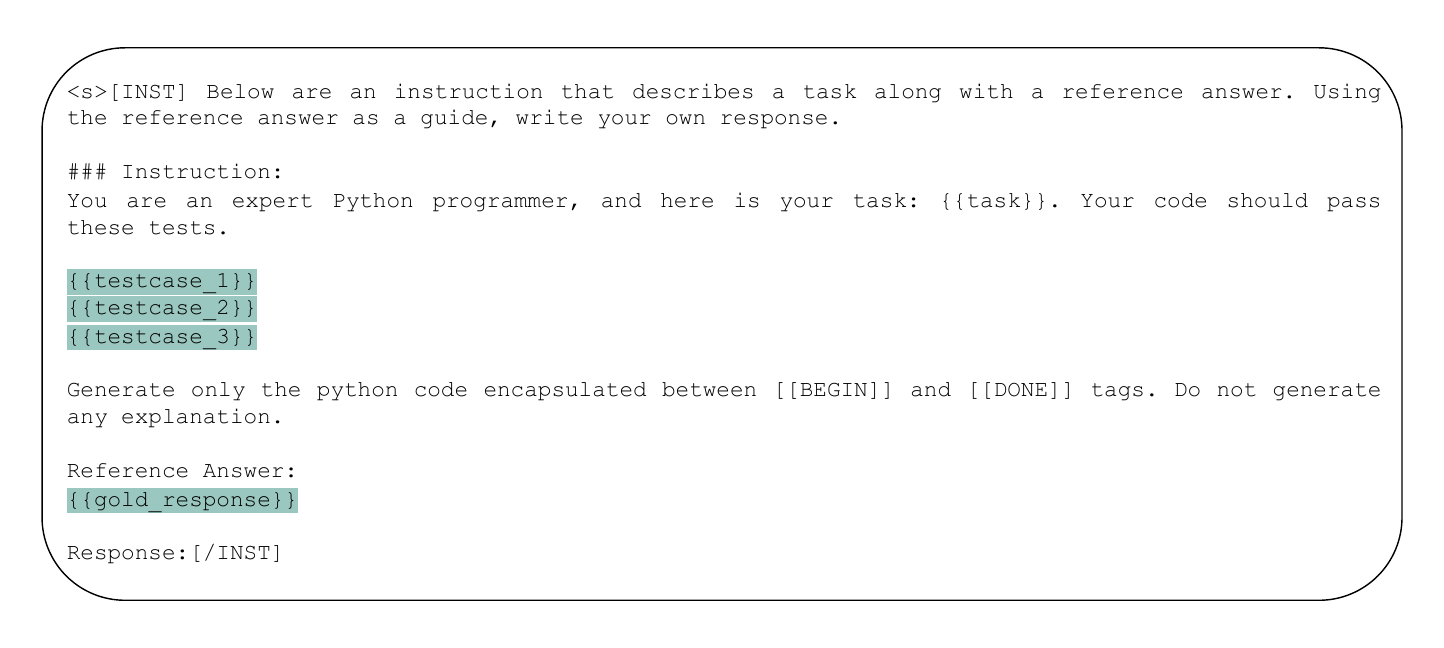}
    \caption{Prompt used for paraphrasing the gold response of training partition of the MBPP dataset.}
    \label{fig:mbpp_gold_para}
\end{figure*}

\begin{figure*}
    \centering
    \includegraphics[width=\linewidth]{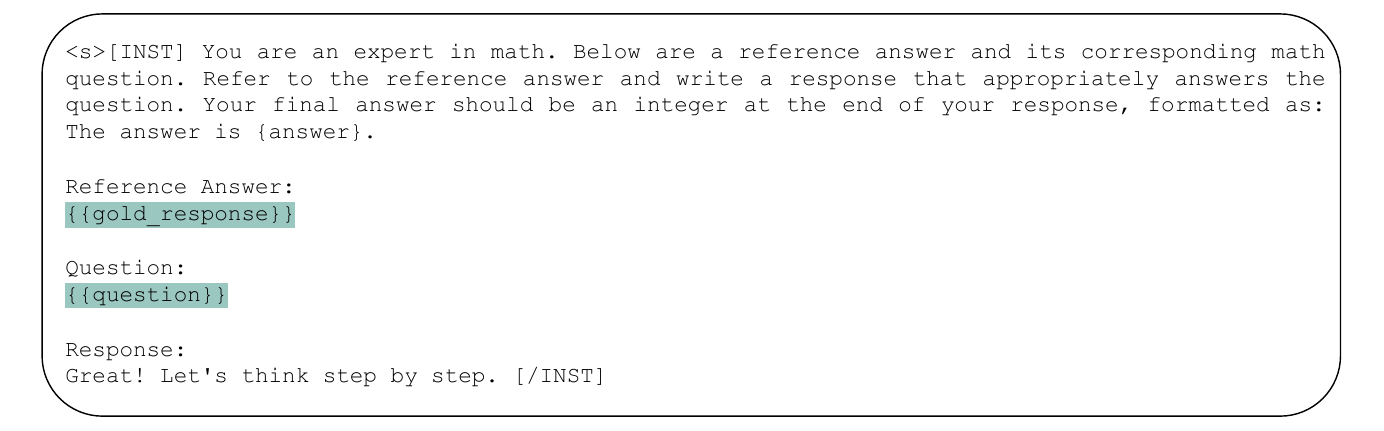}
    \caption{Prompt used for paraphrasing the gold response of training partition of the GSM8K dataset.}
    \label{fig:gsm8k_gold_para}
\end{figure*}

\begin{figure}
    \centering
    \includegraphics[width=\linewidth]{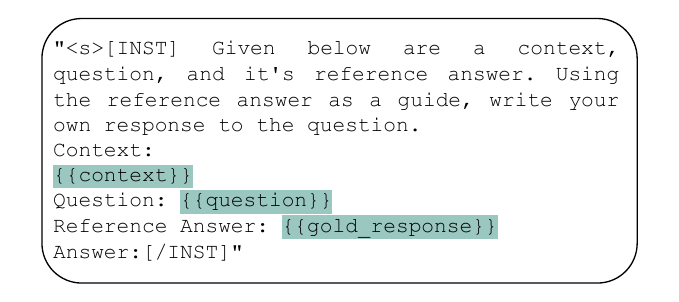}
    \caption{Prompt used for paraphrasing the gold response of training partition of the NQ dataset.}
    \label{fig:nq_gold_para_prompt}
\end{figure}

\begin{figure}
    \centering
    \includegraphics[width=\linewidth]{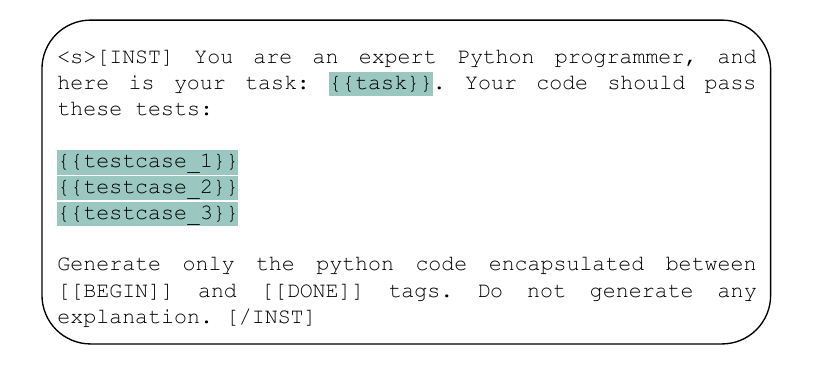}
    \caption{Prompt used for predicting the base model response on MBPP training dataset.}
    \label{fig:mbpp_predict}
\end{figure}

In this section, we list the prompts used with mistral-instruct-v2 to generate the base model responses, gold paraphrases, and training. For the sake of consistency and fair comparison, the same prompts are used for fine-tuning using SFT, SDFT and S3FT techniques. Figure \ref{fig:mbpp_predict} and Figure  \ref{fig:mbpp_gold_para} present the prompts used for generating the base model prediction and gold paraphrasing for MBPP dataset. To simplify the process of extracting the code in the generated output, we always ask the model to generate only the Python code starting with the string "[[BEGIN]]" and ending in the string "[[DONE]]". For the MBPP dataset, the training prompt is the same as the base model's prediction prompt. Figure \ref{fig:gsm8k_base_res_prompt} and Figure \ref{fig:gsm8k_gold_para} present the prompt used for generating the base model response and paraphrasing the gold response. The details of the training prompt are provided in Figure \ref{fig:gsm8k_train_prompt}. Similarly, Figures \ref{fig:nq_base_prompt} and \ref{fig:nq_gold_para_prompt} show the prompts used for generating the base model output and gold paraphrasing, respectively.
\subsection{Datasets Details}
\label{app:data_detail}
\paragraph{GSM8K:} GSM8K is a math word problem dataset comprising of 7473 training examples and 1319 test samples. Each question takes approximately 2-8 steps to solve. Solving these problem require knowledge of basic algebra only. Solutions are human written containing steps in natural language as well. We note that GSM8K does not have a validation set, so we took randomly sampled 150 samples from the dataset as validation.

\paragraph{MBPP:} MBPP stands for Mostly basic python programming dataset. As the name suggests, each example of this dataset contains a task description along with three test cases that the code solving the given task should pass. The gold response contains the python code. There are 372 training examples, 90 validation examples and 500 examples in the testing dataset.

\paragraph{NQ:} NQ is a content-grounded QA dataset. To increase the complexity of the task, \cite{slobodkin2023curious} augment the NQ dataset with unanswerable queries. Here, the grounding content consists of a single paragraph and the gold answers are short phrases.

\begin{figure*}
    \centering
    \includegraphics[width=\linewidth]{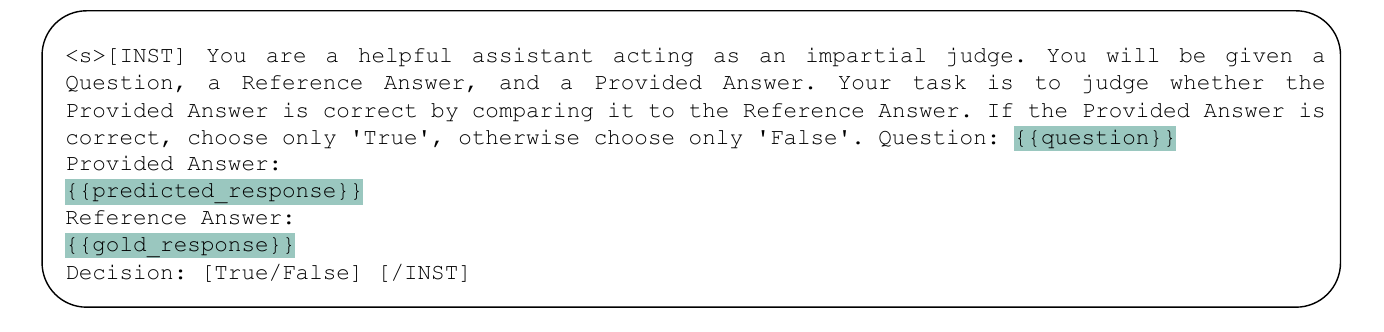}
    \caption{The prompt used for judging the correctness of the responses generated by the model for NQ dataset. Here, {{question}} refer to the question that the model is answering, {{predicted response}} is the response generated by the fine-tuned model and gold response is the gold answer for the question.}
    \label{fig:llm-judge-prompt}
\end{figure*}

\subsection{Training Details}
\label{app:train_detail}
Training for all the experiments was carried out on a single A100 (80 GB) GPU. None of the experiments took more than 1.5 hours to train. To generate the base model's responses, we deployed \mistral using vLLM on a single A100 (80 GB) GPU. It took 1 hour to generate the base model's predictions for GSM8K, 30 minutes to generate the base model's responses for NQ dataset and 15 minutes on MBPP dataset. The entire life cycle, including training data generation, fine-tuning and evaluation, did not take more than 5 hours.

We use a learning rate of $1 \times e^{-4}$. For GSM8K and NQ, we train all the models for 5000 steps, validate after every 500 steps, and select the best checkpoint. For MBPP, we train the models for 1000 steps, validate after every 100 steps, and select the base checkpoint based on the accuracy over the validation set.

For evaluating the GSM8K dataset, we matched the last number of the gold response with the last number extracted from the predicted response. If these answers matched, we considered the generated response to be correct. For evaluating the Python codes, we use bigcode-evaluation-harness \cite{bigcode-evaluation-harness}. LLM-judge's output was parsed to check the correctness of the answers for the NQ dataset. If the judge responded "TRUE", the answer was considered correct; else, if the judge predicted "FALSE", the answer was deemed incorrect. We always used greedy decoding when generating the model responses. Thus, we do a single run of the evaluation and report the numbers. To evaluate the trained models on various benchmarks, the widely popular lm-evaluation-harness \cite{eval-harness} repository was used.

\subsection{S3FT Algorithm}
\label{app:algo}
Here we show the algorithm for S3FT.

\begin{algorithm}
\caption{Training Method}
\begin{algorithmic}[1]
\STATE \textbf{Input:} Base Model $\model_{\theta_0}$ \\
Data $D = \{(\inp_i, \outp_i)\}_{i=1}^{n}$
\STATE $D' \gets \{\}$
\FOR{each example $(\inp_i, \outp_i) \in D$}
    \STATE $\pred_i \gets \model_{\theta_0}(\inp_i)$
    \IF{$\pred_i = \outp_i$}
        \STATE $D' \gets D' \cup \{(\inp_i, \pred_i)\}$
    \ELSE
        \STATE $\tilde{y_i}  \gets \model_{\theta_0}(\outp_i)$
        \IF{$\tilde{y_i}= \outp_i$}
            \STATE $D' \gets D' \cup \{(\inp_i, \tilde{y_i})\}$
        \ELSE
            \STATE $D' \gets D' \cup \{(\inp_i, \outp_i)\}$
        \ENDIF
    \ENDIF
\ENDFOR
\STATE Train $\model_{\theta_0}$ on $D'$ to obtain updated parameters $\theta$
\STATE \textbf{Output:} Updated model parameters $\theta$
\end{algorithmic}
\end{algorithm}

\subsection{Judges for evaluating the model performance}
\label{app:llm-judge}
For tasks like reading comprehension, conventional metrics like BLEU and ROUGE are helpful but inadequate to capture the semantics of the generated responses. Thus, we employ a Language Model (LLM) as an evaluator. Specifically, inspired by \cite{badshah2408reference}, we utilize Mistral-instruct-v2 (7B) model as a judge guided by the prompt illustrated in Figure \ref{fig:llm-judge-prompt}.

\end{document}